\documentclass[a4paper,openany,10pt]{article} 
\usepackage{a4wide} 

\usepackage[backend=biber,style=numeric-comp,citestyle=numeric,sorting=nty,natbib,maxbibnames=10]{biblatex}

\usepackage{graphicx}
\usepackage{tikz}
\usepackage{pgfplots}
\pgfplotsset{compat=1.6}
\usepgfplotslibrary{groupplots}
\usepackage{pgf}
\usetikzlibrary{positioning,shapes,arrows,fit,calc,automata,perspective,3d}
\usepackage{neuralnetwork}
\usepackage{tikz-network}

\usepackage[english]{babel}

\usepackage{a4wide}

\usepackage{amsmath, amsthm, amsfonts, amssymb}
\usepackage{mathabx}   
\usepackage[makeroom]{cancel}    
\usepackage{tipa}      
\usepackage{textcomp}  
\usepackage{latexsym}  
\usepackage{mathrsfs}  

\usepackage{enumitem}

\usepackage{hyperref}
\hypersetup{
	colorlinks=true,       
	linkcolor=blue,
	citecolor=red,
	filecolor=magenta,      
	urlcolor=cyan           
}

\usepackage{cleveref}

\usepackage{hhline}

\usepackage{multirow}
\usepackage{colortbl} 

\usepackage{algpseudocode}
\usepackage{algorithm}

\usepackage{cancel}

\usepackage{subcaption} 

\usepackage{xstring}

\usepackage{ifthen}

\newcommand{\TENS}[1]{\mathbf{#1}} 
\renewcommand{\v}{\ensuremath{\boldsymbol}}				

\newcommand{\N}{\ensuremath{\mathbb{N}}}				
\newcommand{\R}{\ensuremath{\mathbb{R}}}				

\newcommand{\pe}[1]{^{(#1)}}	

\newtheorem{defin}{Definition}[section]

\theoremstyle{definition}

\newtheorem{rem}{Remark}[section]

\bibliography{
References/DellaPapers,References/aiPapers,References/otherPapers} 

\title{
Sparse Implementation of Versatile Graph-Informed Layers
}
\author{
Francesco Della Santa \thanks{Dipartimento di Scienze Matematiche, Politecnico di Torino, Turin, Italy} \thanks{Memeber of the INdAM GNCS Research Group, Rome, Italy}
}

\date{} 

\begin{document}

\maketitle


\begin{abstract}
Graph Neural Networks (GNNs) have emerged as effective tools for learning tasks on graph-structured data. Recently, Graph-Informed (GI) layers were introduced to address regression tasks on graph nodes, extending their applicability beyond classic GNNs. However, existing implementations of GI layers lack efficiency due to dense memory allocation. This paper presents a sparse implementation of GI layers, leveraging the sparsity of adjacency matrices to reduce memory usage significantly. Additionally, a versatile general form of GI layers is introduced, enabling their application to subsets of graph nodes. The proposed sparse implementation improves the concrete computational efficiency and scalability of the GI layers, permitting to build deeper Graph-Informed Neural Networks (GINNs) and facilitating their scalability to larger graphs.


\end{abstract}

\section{Introduction}\label{sec:intro}

Graph Neural Networks (GNNs) are well known as powerful tools for learning tasks on graph-structured data \cite{GNNsurvey2020}, such as semi-supervised node classification, link prediction, and graph classification, with their origin that dates back to the late 2000s \cite{firstGNN_Gori2005,firstGNN_Micheli2009,firstGNN_Scarselli2009}. Recently, a new type of layer for GNNs called Graph-Informed (GI) layer \cite{GINN} has been developed, specifically designed for regression tasks on graph-nodes; indeed, this type of task is not suitable for classic GNNs and, therefore, typically it is approached using MLPs, that do not exploit the graph structure of the data. Nonetheless, the usage of GI layers has been recently extended also to supervised classification tasks (see \cite{dellasanta2024graphinformed}).

The main advantages of the GI layers is the possibility to build Neural Networks (NNs), called Graph-Informed NNs (GINNs), suitable for large graphs and deep architectures. Their good performances, especially if compared with respect to classic MLPs, are illustrated both in \cite{GINN} (regression tasks) and \cite{dellasanta2024graphinformed} (classification task for discontinuity detection).

However, at the time this work is written, existing GI layer implementations have one main limitation. Specifically, all the variables in the codes do not exploit the sparsity of the adjacency matrix of the graph. Therefore, the memory allocated to store a GI layer is much larger, because all the zeros (representing missing connections between graph nodes) are stored even if they are not involved in the GI layer's graph-convolution operation. The problem of this ``dense'' implementation is that the computer's memory can be easily saturated, especially when building GINNs based on large graphs and/or many GI layers (i.e., deep architectures). Therefore, the principal scope of this work is to present a new implementation of the GI layers that is sparse; i.e., an implementation that exploits the sparsity of the adjacency graph, reducing concretely the memory allocated for storing GI layers and GINNs in general.

In \cite{GINN}, the definition of a GI layer is very general and can be easily applied to any kind of graph that is directed or not and that is weighted or not. However, the original definition can be further generalized, extending the action of GI layers to subsets of graph nodes, introducing few more details. In this work, we introduce such a kind of generalization, called \emph{versatile general form} of GI layers.

Summarizing, the proposed implementation of GI layers that is illustrated in this work allows the handling of sub-graphs and is optimized to leverage the sparse nature of adjacency matrices, resulting in significant improvements in computational efficiency and memory utilization. In particular, the sparse implementation enables the construction of deeper GINNs and facilitates scalability for larger graphs.

The work is organized as follows: in \Cref{sec:GIlayers} the GI layers are formally introduced, recalling their inner mechanisms. In \Cref{sec:GIlayer_versatile} the versatile general form of GI layers is formally defined, explaining similarities and differences with respect to the previous version of GI layers. In \Cref{sec:sparseGI_tf} we describe in details the sparse implementation of the versatile GI layers, both reporting the pseudocode (\Cref{sec:pseudocode}) and the documentation of the python class available in the public repository (\url{https://github.com/Fra0013To/GINN}, march 2024 update). In \Cref{sec:conclusions} we summarize the results of the work.

\section{Graph-Informed Layers}\label{sec:GIlayers}

Graph-Informed (GI) Layers \cite{GINN} are NN layers based on a new variant of the graph-convolution operation. Their simplest version consists of one input feature per node and one output feature per node; in this case, the action of a GI layer is based on the adjacency matrix $A\in\R^{n\times n}$ of a given graph (without self-loops) and it is a function $\mathcal{L}^{GI}:\R^n\rightarrow\R^n$ such that
\begin{equation}\label{eq:GI_action_simple}
    \mathcal{L}^{GI}(\v{x}) 
    = \v{\sigma}\left( \widehat{W}^T \, \v{x} + \v{b}\right) = 
    \v{\sigma}\left( (\mathrm{diag}(\v{w}) (A + \mathbb{I}_n))^T\, \v{x} + \v{b}\right)
    \,,
\end{equation}
for each vector of input features $\v{x}\in\R^n$, where:
\begin{itemize}
    \item $\v{w}\in\R^n$ is the vector containing the $n$ layer weights, associated to the $N$ graph nodes and $\widehat{W}$ denotes the product $\mathrm{diag}(\v{w}) (A + \mathbb{I}_n)$;

    \item $\v{\sigma}:\R^n\rightarrow\R^n$ is the element-wise application of the activation function $\sigma$;
    
    \item $\v{b}\in\R^n$ is the bias vector.
\end{itemize}

In brief, \eqref{eq:GI_action_simple} is equivalent to the action of a Fully-Connected (FC) layer where the weights are shared and constrained such that
\begin{equation*}\label{eq:GI_weights_simple}
    w_{ij}=
    \begin{cases}
    w_i\,,\quad & \text{if }a_{ij}\neq 0 \text{ or }i=j\\
    0\,,\quad & \text{otherwise}
    \end{cases}
    \,.
\end{equation*}

Layers characterized by \eqref{eq:GI_action_simple} can be generalized to receive any arbitrary number $K\geq 1$ of input features per node and to return any arbitrary number $F\geq 1$ of output features per node. Then, the general definition of a GI layer is the following.

\begin{figure}[htb]
	\centering
        \resizebox{0.35\textwidth}{!}{
	\begin{tikzpicture}[3d view = {15}{15}]
	\draw[-{Latex[scale = 0.7]}] (-4, -4, 0) -- (-3, -4, 0) node[below] {\footnotesize{col.-dim.}}; 
	\draw[-{Latex[scale = 0.7]}] (-4, -4, 0) -- (-4, -3, 0) node[above] {\footnotesize{$3^{rd}$ dim.}}; 
	\draw[-{Latex[scale = 0.7]}] (-4, -4, 0) -- (-4, -4, -1) node[below] {\footnotesize{row-dim.}}; 
	
	%
	\begin{scope}[canvas is yz plane at x = -2]
		\draw[opacity=0.5,fill=lightgray] (1.5, -1.5) rectangle (-4.5, 1.5);
		\node [] (W11) at (-0.85,-0.25) {$\widehat{W}^{(1,1)}$};
        
	\end{scope}
	\begin{scope}[canvas is yz plane at x = -2]
		\node [] (W1k) at (-1,-2) {$\vdots$};
	
	\end{scope}
	\begin{scope}[canvas is yz plane at x = -2]
	\draw[opacity=0.5,fill=lightgray] (1.5, -2.5) rectangle (-4.5, -5.5);
	\node [] (W1K) at (-0.9,-4.25) {$\widehat{W}^{(K,1)}$};
	
	\end{scope}
	\begin{scope}[canvas is yz plane at x = -2]
	\draw[red] (1.5, -5.5) rectangle (-4.5, 1.5);
	\node [] (W1) at (-1,2.25) {\color{red}$\widetilde{W}^{(1)}$};
	
	\end{scope}
    
    \begin{scope}[canvas is yx plane at z = 1.5]
    	\draw[opacity=0.5] (1.5, -2) rectangle (-4.5, 3.5);
	\end{scope} 
	\node[opacity=0.5] (W2_WFm1) at (1,-0.9,1.5) {\footnotesize{$\widetilde{W}^{(2)},\ldots ,\widetilde{W}^{(F-1)}$}};
	
	\begin{scope}[canvas is yx plane at z = -5.5]
	\draw[opacity=0.5] (1.5, -2) rectangle (-4.5, 3.5);
	
	\end{scope} 

	\begin{scope}[canvas is yz plane at x = 3.5]
	\draw[opacity=0.5,fill=lightgray] (1.5, -1.5) rectangle (-4.5, 1.5);
	\node [] (WF1) at (-0.85,-0.25) {$\widehat{W}^{(1,F)}$};
	
	\end{scope}
	\begin{scope}[canvas is yz plane at x = 3.5]
	\node [] (WFk) at (-1,-2) {$\vdots$};
	
	\end{scope}
	\begin{scope}[canvas is yz plane at x = 3.5]
	\draw[opacity=0.5,fill=lightgray] (1.5, -2.5) rectangle (-4.5, -5.5);
	\node [] (WFK) at (-0.9,-4.25) {$\widehat{W}^{(K,F)}$};
	
	\end{scope}
	\begin{scope}[canvas is yz plane at x = 3.5]
	\draw[blue] (1.5, -5.5) rectangle (-4.5, 1.5);
	\node [] (WF) at (-1,2.25) {\color{blue}$\widetilde{W}^{(F)}$};
	
	\end{scope}

\end{tikzpicture}
        }
	\caption{Tensor $\widetilde{\TENS{W}}$ obtained concatenating along the second dimension the matrices $\widetilde{W}^{(1)},\ldots , \widetilde{W}^{(F)} \in\R^{nK \times n}$. Before the concatenation, the matrices are reshaped as tensors in $\R^{nK\times 1\times n}$.}
	\label{fig:tensor_What_concatenation_3d}
\end{figure}
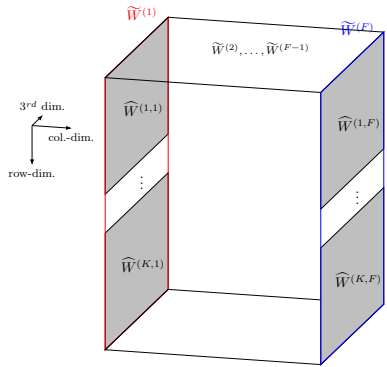

\begin{defin}[Graph Informed Layer - General form \cite{GINN}]\label{def:GIlayer_general}
    Let $A\in\R^{n\times n}$ be the adjacency matrix of a given graph. Then, a \emph{GI layer with $K\in\N$ input features and $F\in\N$ output features} is a NN layer with $n F$ units connected to a layer with outputs in $\R^{n\times K}$ and having a characterizing function $\mathcal{L}^{GI}:\R^{n\times K}\rightarrow\R^{n\times F}$ defined by
    \begin{equation}\label{eq:GIlayer_general}
    	\mathcal{L}^{GI}(X) = \v{\sigma}\left( \widetilde{\TENS{W}}^T \mathrm{vertcat}(X) + B \right)\,,
    \end{equation}
    where:
    \begin{itemize}
        \item $X\in\R^{n\times K}$ is the input matrix (i.e., the output of the previous layer) and $\mathrm{vertcat}(X)$ denotes the vector in $\R^{nK}$ obtained concatenating the columns of $X$;
        
        \item the tensor $\widetilde{\TENS{W}}\in \R^{nK\times F\times n}$ is defined as the concatenation along the second dimension (i.e., the column-dimension) of the matrices $\widetilde{W}^{(1)},\ldots ,\widetilde{W}^{(F)}$, such that
        \begin{equation}\label{eq:Wtilde_filter_concat}
        	\widetilde{W}^{(l)} := 
        	\begin{bmatrix}
        		\widehat{W}\pe{1,l}\\
        		\vdots\\
        		\widehat{W}\pe{K, l}
        	\end{bmatrix} 
        	=
        	\begin{bmatrix}
        		\mathrm{diag}(\v{w}\pe{1, l})(A+\mathbb{I}_n)\\
        		\vdots\\
        		\mathrm{diag}(\v{w}\pe{K, l}(A+\mathbb{I}_n)
        	\end{bmatrix}
        	\in\R^{nK\times n}\,,
        \end{equation}
        for each $l=1,\ldots ,F$, and where $\v{w}\pe{k,l}\in\R^n$ is the weight vector characterizing the contribute of the $k$-th input feature to the computation of the $l$-th output feature of the nodes, for each $k=1,\ldots ,K$, and $l=1,\ldots ,F$. 
        
        Before the concatenation, the matrices $\widetilde{W}^{(1)},\ldots ,\widetilde{W}^{(F)}$ are reshaped as tensors in $\R^{nK\times 1\times n}$ (see \Cref{fig:tensor_What_concatenation_3d});
        
        \item the operation $\widetilde{\TENS{W}}^T \mathrm{vertcat}(X)$ is a tensor-vector product;
        
        \item $B\in\R^{n\times F}$ is the matrix of the biases, i.e., each column $\v{b}_{\cdot l}$ is the bias vector corresponding to the $l$-th output feature of the nodes.
    \end{itemize}
\end{defin}

\noindent Additionally, pooling and mask operations can be added to GI layers (see \cite{GINN} for more details). A NN made of GI layers is called \emph{Graph-Informed Neural Network} (GINN) \cite{GINN}.

We point the attention of the reader to the fact that the tensor $\widetilde{\TENS{W}}$ typically is sparse. Then, it is important to exploit the sparsity of this tensor for reducing the memory cost of the GINN.

\section{Versatile GI Layers}\label{sec:GIlayer_versatile}

We observe that the GI layer definition is very general and can be easily extended to submatrices of $A$, focusing the GI layer on connections between two subsets $V_1$ and $V_2$ of nodes in the graph; moreover, we can also rescale the action of the graph-convolution self-loops (i.e., the connections of node $v_i$ with itself in the graph-convolution operation) multiplying $\mathbb{I}_n$ in \eqref{eq:Wtilde_filter_concat} by a scalar value $\lambda\in\R$. 
Under these circumstances, we can extend \Cref{def:GIlayer_general} as follows.

\begin{defin}[Graph Informed Layer - Versatile general form]\label{def:GIlayer_versatile}
    Let $A\in\R^{n\times n}$ be the adjacency matrix of a given graph $G=(V, E)$ and let $V_1,V_2\subseteq V$ be two subsets of $n_1,n_2\in\N$ nodes, respectively. Then, a \emph{GI layer with $K\in\N$ input features, $F\in\N$ output features, and related to the connections of nodes $V_1$ with nodes $V_2$} is a NN layer with $n_2 F$ units connected to a layer with outputs in $\R^{n_1\times K}$ and having a characterizing function $\mathcal{L}^{GI}:\R^{n_1\times K}\rightarrow\R^{n_2\times F}$ defined by \eqref{eq:GIlayer_general}
    where the equation \eqref{eq:Wtilde_filter_concat} defyining the tensor $\widetilde{\TENS{W}}\in\R^{n_1K\times F\times n_2}$ changes into 
        \begin{equation}\label{eq:versatileWtilde_filter_concat}
        	\widetilde{W}^{(l)} := 
        	\begin{bmatrix}
        		\widehat{W}\pe{1, l}\\
        		\vdots\\
        		\widehat{W}\pe{K, l}
        	\end{bmatrix} 
        	=
        	\begin{bmatrix}
        		\mathrm{diag}(\v{w}\pe{1, l})\left .(A+\lambda\mathbb{I}_n)\right|_{V_1,V_2}\\
        		\vdots\\
        		\mathrm{diag}(\v{w}\pe{K, l})\left .(A+\lambda\mathbb{I}_n)\right|_{V_1,V_2}
        	\end{bmatrix}
        	\in\R^{n_1K\times n_2}\,,
        \end{equation}
        with $\v{w}\pe{k,l}\in\R^{n_1}$, for each $k=1,\ldots ,K$, and $l=1,\ldots ,F$, and with $\left .(A+\lambda \mathbb{I}_n)\right|_{V_1,V_2}$ that is the $n_1$-by-$n_2$ submatrix of $(A + \lambda \mathbb{I}_n)$ with respect to nodes $V_1$ and $V_2$. Similarly, the bias matrix is now $B\in \R^{n_2\times F}$. The scalar value $\lambda\in\R$ denotes the rescaling factor for the self-loop connections of the graph-convolution.
\end{defin}

The advantages of a versatile GI layer (\Cref{def:GIlayer_versatile}), with respect to a general GI layer (\Cref{def:GIlayer_general}), is the possibility of learning also graph-structured functions $\v{F}:\R^{n_1}\rightarrow \R^{n_2}$ defined on subsets $V_1,V_2\subseteq V$ of graph nodes. Actually, general GI layers can be used for building GINNs able to learn functions like the $\v{F}$ above, but ``neutral'' inputs must be used for the NN's input units corresponding to the graph nodes in $V\setminus V_1$ and the mask operation is necessary to ``focus'' on the NN's outputs corresponding to the graph nodes in $V\setminus V_2$. Of course, a versatile GI layer is equivalent to a general GI layer if $V_1\equiv V_2 \equiv V$ and $\lambda = 1$. 

Concerning the value of $\lambda$, it is suggested to use $\lambda=1$ if the graph $G$ is not weighted; on the other hand, if $G$ is a weighted graph and, therefore, the nonzero values of $A$ are not necessarily equal to $1$ (e.g., see \cite{dellasanta2024graphinformed}), then the value of $\lambda$ can be chosen according to the rule for the edge weights of $G$.

In the next section, we focus on the sparse implementation of versatile GI layers with respect to the Tensorflow (TF) framework \cite{tensorflow2015-whitepaper} and the operations available for sparse arrays.

\section{Sparse Versatile GI Layers in Tensorflow}\label{sec:sparseGI_tf}

The first TF implementation of general GI layers (\Cref{def:GIlayer_general}, \cite{GINN}) assembles the tensor $\widetilde{\TENS{W}}$ as a dense TF tensor and exploits the tensordot operation for a fast and efficient computation of the tensor product $\widetilde{\TENS{W}}^T \widetilde{X}$, where $\widetilde{X}\in\R^{nK\times M}$ denotes a batch of $M\in\N$ concatenated vectors ${\rm vertcat}(X_1),\ldots , {\rm vertcat}(X_M)$ (vectorizations of the input matrices $X_1,\ldots ,X_M$, respectively). Nonetheless, a dense implementation of $\widetilde{\TENS{W}}$ can result in particularly memory expensive GINNs, especially for large graphs ($n\gg 1$), features $K,F > 1$, and large number of GI layers (i.e., deep GINNs). Therefore, a sparse implementation of GI layers is particularly useful. 

Unfortunately, at the time this work is written, a version of the tensordot operation between two sparse tensors is not available yet in the TF's module \emph{tf.sparse} \cite{tf_sparse_mod}; therefore, an easy translation from dense to sparse implementation is not possible. On the other hand, the function \emph{sparse\_dense\_matmul} of the \emph{tf.sparse} module allows to compute the product between a sparse matrix (sparse 2-dimensional TF tensor) and a dense matrix (dense 2-dimensional TF tensor), or vice-versa. 

In this section, we describe the sparse implementation of versatile GI layers (\Cref{def:GIlayer_versatile}) through the \emph{sparse\_dense\_matmul} TF function. Nonetheless, the procedure is generalizable to any other Deep Learning framework endowed with similar functions for sparse arrays (e.g., \cite{torch2011}). In particular, we report the pseudo-code describing the operations performed by the sparse implementation of the new versatile GI layer and the documentation related to the new TF layer class available on the repository, based on this pseudo-code.

\subsection{Pseudocode}\label{sec:pseudocode}

In the following algorithm, we report a pseudocode describing the operations performed by our sparse implementation of versatile GI layers. For the TF implementation (python language), we point the reader to the GINN's repository (\url{https://github.com/Fra0013To/GINN}, march 2024 update).

In particular, \Cref{alg:versGI_action} describes the action of the layer with respect to a generic batch of $M\in\N$ inputs $X_1,\ldots ,X_M\in\R^{n_1\times K}$.

\begin{algorithm}[htb!]
\caption{Versatile Sparse GI Layer's Action}\label{alg:versGI_action}
\begin{algorithmic}[1]
\Require \quad

$A|_{V_1,V_2}\in\R^{n_1\times n_2}$: \emph{sparse} sub-matrix of the adjacency matrix $A$;

$\lambda\in\R$: rescaling factor for the self-loop connections of the graph-convolution.

$\v{w}\pe{k,l}\in\R^{n_1}$: weight vectors, for each $k=1,\ldots ,K$ and $l=1,\ldots ,F$;

$B\in\R^{n_2\times F}$: bias matrix;

$\sigma$: activation function;

$\TENS{X}\in\R^{n_1\times K\times M}$: batch of $M\in\N$ input matrices $X_1,\ldots ,X_M\in\R^{n_1\times K}$, each one storing $K$ input features for nodes in $V_1$.

\Ensure \quad

$\TENS{Y}\in\R^{n_2\times F\times M}$: batch of $M\in\N$ GI layer outputs with respect to $X_1,\ldots ,X_M$; i.e., batch of $M$ matrices of $F$ output features for nodes in $V_2$.

\quad

\State $\widehat{A} \gets (A + \lambda\mathbb{I}_n)|_{V_1,V_2}$

\State $\widetilde{X}\in\R^{n_1K\times M} \gets$ concat. along columns ${\rm vertcat}(X_1),\ldots , {\rm vertcat}(X_M)$; i.e., reshape $\TENS{X}$

\State $\widetilde{Y}_{\rm list} \gets$ empty list

\For{$l=1,\ldots ,F$}
    \State $\widetilde{W}^{(l)}\in\R^{n_1K\times n_2} \gets$ concat. along rows  $\v{w}\pe{1,l} * \widehat{A},\ldots, \v{w}\pe{K,l} * \widehat{A}$ (see \eqref{eq:versatileWtilde_filter_concat}, $*$: row-wise prod.)
    \State $\widetilde{Y}^{(l)}\in\R^{n_2\times M} \gets$ sparse-dense matrix multiplication $(\widetilde{W}^{(l)\,T} \widetilde{X})$
    \State $\widetilde{Y}^{(l)}\in\R^{n_2\times 1\times M} \gets$ reshape $\widetilde{Y}^{(l)}$
    \State add $\widetilde{Y}^{(l)}$ to $\widetilde{Y}_{\rm list}$
\EndFor

\State $\widetilde{\TENS{Y}}\in\R^{n_2\times F\times M} \gets$ concat. along columns the arrays in $\widetilde{Y}_{\rm list}$. 

\State $\TENS{Y}\in\R^{n_2\times F\times M} \gets \v{\sigma}( \widetilde{\TENS{Y}} \oplus B )$ \qquad \qquad \qquad \qquad \qquad \qquad \quad ($\oplus$: vectorized sum, along $3^{rd}$ dim)

\Return{$\TENS{Y}$}

\end{algorithmic}
\end{algorithm}

\begin{rem}[Tensors' shapes - Theory and practice]\label{rem:tens_shapes}
    The pseudo-code illustrated in \Cref{alg:versGI_action} and all the formulas in the previous sections adopt the classic mathematical convention where vectors are considered as column vectors; then, batches of matrices are represented as three-dimensional tensors where the matrices are aligned along the third dimension. On the other hand, almost all the Deep Learning frameworks (TF included) have been developed for working with row vectors; therefore, a batch of matrices is obtained stacking the matrices along the first dimension. For example, the input tensor $\TENS{X}$ of \Cref{alg:versGI_action} is a tensor of shape $n_1$-by-$K$-by-$M$ if we use the mathematical convention, but it is a tensor of shape $M$-by-$n_1$-by-$K$ if we are working with a Deep Learning framework. In the latter case, the operations in \Cref{alg:versGI_action} changes only with respect to the shape of the vectors, matrices, and tensors involved; the meaning and the result of the procedure doesn't change ($\TENS{Y}\in\R^{n_2\times F\times M}$ in mathematical convention, $\TENS{Y}\in\R^{M\times n_2\times F}$ in ``Deep Learning framwork convention'').
\end{rem}

\subsection{Usage Documentation}\label{sec:doc}

The implementation of versatile general GI layers overwrites the previous class \texttt{Graph\-Informed} keras layers in the GINN's repository (\url{https://github.com/Fra0013To/GINN}, march 2024 update). Nonetheless, the previous class in preserved, changing name into \texttt{Dense\-Non\-versatile\-Graph\-Informed}; the new \texttt{GraphInformed} class is obtained as a subclass of the previous one, even if the \texttt{build} and \texttt{call} methods are overwritten.
In this subsection, we report the detailed documentation related to the new \texttt{Graph\-Informed} class. Many initialization arguments points toward the documentation of the \texttt{Dense} layer class of \emph{keras} in TF, because \texttt{GraphInformed} is a sub-class of it (more precisely, \texttt{Dense\-Non\-versatile\-Graph\-Informed} is a sub-class of \texttt{Dense}).

\begin{description}
    \item[Initialization Arguments:] the arguments are listed in the same order of the code in the repository.
    \begin{itemize}
        \item \texttt{adj\_mat}: the matrix $A|_{V_1, V_2}$. It must be a \texttt{sci\-py.\-sparse.\-dok\-\_\-matrix} or \texttt{sci\-py.\-sparse.\-dok\-\_\-array} (see \cite{2020SciPy-NMeth}), or a dictionary describing the adjacency matrix using the following keys:
        \begin{itemize}
            \item \emph{keys}: list of tuples $(i,j)$ denoting the non-zero elements of the matrix $A|_{V_1,V_2}$;
            \item \emph{values}: list of non-zero values of $A|_{V_1,V_2}$, corresponding to \emph{keys};
            \item \emph{rowkeys\_custom}: list of indices $i_1,\ldots ,i_{n_1}$ denoting the nodes in $V_1$. If \texttt{None}, we assume that they are $0,\ldots ,n_1 - 1$;
            \item \emph{colkeys\_custom}: list of indices $j_1,\ldots ,j_{n_2}$ denoting the nodes in $V_2$. If \texttt{None}, we assume that they are $0,\ldots ,n_2 - 1$;
            \item \emph{keys\_custom:} list of tuples $(i_k,j_h)$ that ``translate'' the tuples in \emph{keys} with respect to the indices stored in \emph{rowkeys\_custom} and \emph{colkeys\_custom}. If \texttt{None}, we assume that this list is equal to the one stored in \emph{keys}.
        \end{itemize}
        Such a kind of dictionary can be easily obtained from a sparse matrix using the \texttt{sparse2dict} function defined in a module of the repository.
        \item \texttt{rowkeys}: list, default \texttt{None}. List containing the indices of the nodes in $V_1$. If \texttt{None}, we assume that the indices are $0, \ldots ,n_1 - 1$. Any list is automatically sorted in ascending order. This argument is ignored if the \texttt{adj\_mat} argument is a dictionary.
        \item \texttt{colkeys}: list, default \texttt{None}. List containing the indices of the nodes in $V_2$. If \texttt{None}, we assume that the indices are $0, \ldots ,n_2 - 1$. Any list is automatically sorted in ascending order. This argument is ignored if the \texttt{adj\_mat} argument is a dictionary.
        \item \texttt{selfloop\_value}: float, default \texttt{1.0}. Rescaling factor of the graph-convolution self-loops, i.e. the value $\lambda\in\R$ in $(A + \lambda \mathbb{I}_n)|_{V_1, V_2}$.
        \item \texttt{num\_filters}: integer, default \texttt{1}. Integer value describing the number $F\in\N$ of filters (i.e., output features per node) of the layer (see \Cref{def:GIlayer_versatile}). The value $K\in\N$ of input features per node is inferred directly from the inputs.
        \item \texttt{activation}, \texttt{use\_bias}, \texttt{kernel\_initializer}, \texttt{bias\_initializer}, \texttt{kernel\_regularizer}, \texttt{bias\_regularizer}, \texttt{activity\_regularizer}, \texttt{kernel\_constraint}, \texttt{bias\_constraint}: see the \texttt{tensorflow.keras.layers.Dense} class;
        \item \texttt{pool}: string, default \texttt{None}. String describing a ``reducing function'' of TF (e.g., \texttt{'re\-du\-ce\-\_\-mean'}, \texttt{'re\-du\-ce\-\_\-max'}, etc.);
        \item \texttt{**options}: keyword arguments, see the \texttt{tensorflow.keras.layers.Dense} class.
    \end{itemize}
    
    \item[Attributes:] we exclude from this list the attributes inherited from \texttt{ten\-sor\-flow.\-ke\-ras.\-layers.\-Dense}, unless they have been specifically re-defined for implementation purposes. The attributes are listed in alphabetical order.
    \begin{itemize}
        \item \texttt{adj\_mat}: dictionary describing the matrix $(A+\lambda\mathbb{I}_n)|_{V_1,V_2}$;
        \item \texttt{adj\_mat\_original}: dictionary describing the matrix $A|_{V_1,V_2}$ (see the \texttt{adj\_mat} initialization argument above);
        \item \texttt{adj\_mat\_tf}: sparse TF tensor describing the matrix $(A+\lambda\mathbb{I}_n)|_{V_1,V_2}$;
        \item \texttt{bias}: \emph{trainable} TF variable, TF tensor $\TENS{B}\in\R^{1\times n_2 \times F}$, where $n_2=|V_2|$ and $F\in\N$ is the number of output features per node (see \texttt{num\_filters} below). Tensor $\TENS{B}$ is just the reshaped bias matrix $B\in\R^{n_2\times F}$.
        \item \texttt{kernel}: \emph{trainable} TF variable, TF tensor $\TENS{W}\in\R^{n_1K\times 1 \times F}$, where $n_1=|V_1|$, $K\in\N$ is the number of input features per node (see \texttt{num\_features} below), and $F\in\N$ is the number of output features per node (see \texttt{num\_filters} below). Tensor $\TENS{W}$ is a reshaped matrix $W\in\R^{n_1 K\times F}$ with columns the concatenation of the weight vectors $\v{w}\pe{1,l},\ldots ,\v{w}\pe{K,l}$, foe each $l=1,\ldots ,F$.
        \item \texttt{num\_features}: integer $K\in\N$ of input features per node. Inferred by layer inputs during model creation.
        \item \texttt{num\_filters}: integer $F\in\N$ of output features per node. Defined during layer initialization.
        \item \texttt{pool}: ``reducing function'' of TF, inferred from the initialization argument \texttt{pool}. If this argument is \texttt{None}, also the attribute is \texttt{None}.
        \item \texttt{pool}\_str: value assigned to the initialization argument \texttt{pool} during the object creation.
    \end{itemize}
    
    \item[Methods:] we exclude from this list the methods inherited from \texttt{ten\-sor\-flow.\-ke\-ras.\-layers.\-Dense}, unless they have been specifically re-defined for implementation purposes. The methods are listed in alphabetical order.
    \begin{itemize}
        \item \texttt{build(self, input\_shape)}: method for building the layer. We overwrite the one inherited from \texttt{ten\-sor\-flow.\-ke\-ras.\-layers.\-Dense}. This method initialize the attributes \texttt{num\_features} and \texttt{adj\_mat\_tf}, and also the trainable parameters stored in the attributes \texttt{bias} and \texttt{kernel}.
        \item \texttt{call(self, input)}: this method describe the action of the layer; i.e., given an input tensor \texttt{input}$=\TENS{X}\in\R^{M\times n_1\times K}$, equivalent to a batch of $M\in\N$ input matrices $X_1,\ldots ,X_M\in\R^{n_1\times K}$, the method performs the operations described in \Cref{alg:versGI_action}, returning a tensor $\TENS{Y}\in\R^{M\times n_2\times F}$. Concerning the shape of the input and output tensors, we recall \Cref{rem:tens_shapes}.
    \end{itemize}
\end{description}

\section{Conclusion}\label{sec:conclusions}

In this work, we introduced a further generalization of the Graph-Informed layers and a novel, improved, tensorflow implementation of them.

Specifically, we defined the so-called versatile GI layers (\Cref{def:GIlayer_versatile}), extending the previous definition to work with submatrices $A|_{V_1,V_2}$ of the adjacency matrix and adding the possibility to rescale the self-loops added by the graph-convolution operation. From the practical point of view, this new GI layer definition permits to work with subgraphs and makes the mask operation obsolete. 

Concerning the novel and improved tensorflow implementation of the new GI layers, it exploits the sparse nature of the adjacency matrices. This method grants notable advantages, first of all the possibility to build larger and deeper Graph-Informed NNs. Indeed, the memory allocated for the new implementation of the GI layers strictly depends on the nonzero elements in $A|_{V_1,V_2}$, avoiding to store the (possibly many) zero elements that occur during the computations of the outputs.




% ...





\section*{Acknowledgements}
This study was carried out within the FAIR - Future Artificial Intelligence Research and received funding from the European Union Next-GenerationEU (PIANO NAZIONALE DI RIPRESA E RESILIENZA (PNRR) – MISSIONE 4 COMPONENTE 2, INVESTIMENTO 1.3 – D.D. 1555 11/10/2022, PE00000013). This manuscript reflects only the authors’ views and opinions, neither the European Union nor the European Commission can be considered responsible for them.
The research has also been partially supported by INdAM-GNCS.


\printbibliography

@article{GINN, 
	author = {Berrone, Stefano and {Della Santa}, Francesco and Mastropietro, Antonio and Pieraccini, Sandra and Vaccarino, Francesco}, 
	title = {Graph-Informed Neural Networks for Regressions on Graph-Structured Data},
	journal = {Mathematics},
	doi = {10.3390/math10050786}, 
	url = {https://doi.org/10.3390%2Fmath10050786}, 
	year = {2022},
	month = {mar}, 
	publisher = {{MDPI} {AG}}, 
	volume = {10}, 
	number = {5}, 
	pages = {786}, 
}

@misc{dellasanta2024graphinformed,
      title={Graph-Informed Neural Networks for Sparse Grid-Based Discontinuity Detectors}, 
      author={Francesco {Della Santa} and Sandra Pieraccini},
      year={2024},
      eprint={2401.13652},
      archivePrefix={arXiv},
      primaryClass={cs.LG}
}

@article{GNNsurvey2020,
	author={Wu, Zonghan and Pan, Shirui and Chen, Fengwen and Long, Guodong and Zhang, Chengqi and Yu, Philip S.},
	journal={IEEE Transactions on Neural Networks and Learning Systems}, 
	title={A Comprehensive Survey on Graph Neural Networks}, 
	year={2021},
	volume={32},
	number={1},
	pages={4-24},
	doi={10.1109/TNNLS.2020.2978386}
}

@INPROCEEDINGS{firstGNN_Gori2005,
	author={Gori, M. and Monfardini, G. and Scarselli, F.},
	booktitle={Proceedings. 2005 IEEE International Joint Conference on Neural Networks, 2005.}, 
	title={A new model for learning in graph domains}, 
	year={2005},
	volume={2},
	number={},
	pages={729-734 vol. 2},
	doi={10.1109/IJCNN.2005.1555942}
}

@ARTICLE{firstGNN_Micheli2009,
	author={Micheli, Alessio},
	journal={IEEE Transactions on Neural Networks}, 
	title={Neural Network for Graphs: A Contextual Constructive Approach}, 
	year={2009},
	volume={20},
	number={3},
	pages={498-511},
	doi={10.1109/TNN.2008.2010350}
}

@ARTICLE{firstGNN_Scarselli2009,
	author={Scarselli, Franco and Gori, Marco and Tsoi, Ah Chung and Hagenbuchner, Markus and Monfardini, Gabriele},
	journal={IEEE Transactions on Neural Networks}, 
	title={The Graph Neural Network Model}, 
	year={2009},
	volume={20},
	number={1},
	pages={61-80},
	doi={10.1109/TNN.2008.2005605}
}

@misc{tensorflow2015-whitepaper,
	title={ {TensorFlow}: Large-Scale Machine Learning on Heterogeneous Systems},
	url={https://www.tensorflow.org/},
	note={Software available from tensorflow.org},
	author={Mart\'{i}n~Abadi and Ashish~Agarwal and Paul~Barham and Eugene~Brevdo and Zhifeng~Chen and Craig~Citro and Greg~S.~Corrado and
	Andy~Davis and Jeffrey~Dean and	Matthieu~Devin and Sanjay~Ghemawat and Ian~Goodfellow and Andrew~Harp and Geoffrey~Irving and Michael~Isard and Yangqing Jia and Rafal~Jozefowicz and Lukasz~Kaiser and Manjunath~Kudlur and Josh~Levenberg and Dandelion~Man\'{e} and Rajat~Monga and Sherry~Moore and Derek~Murray and Chris~Olah and Mike~Schuster and Jonathon~Shlens and Benoit~Steiner and Ilya~Sutskever and Kunal~Talwar and Paul~Tucker and Vincent~Vanhoucke and Vijay~Vasudevan and Fernanda~Vi\'{e}gas and Oriol~Vinyals and Pete~Warden and Martin~Wattenberg and Martin~Wicke and Yuan~Yu and Xiaoqiang~Zheng},
	year={2015},
}

@manual{tf_sparse_mod,
	title={Tensorflow - Module: tf.sparse},
	author={},
	note={(Accessed on January 2024)},
	url={https://www.tensorflow.org/api_docs/python/tf/sparse}
}

@inproceedings{torch2011,
	author={Collobert, Ronan and Kavukcuoglu, Koray and Farabet, Cl\`{e}ment},
	booktitle = {BigLearn, NIPS Workshop, number EPFL-CONF-192376},
	pages = {},
	title={Torch7: AMatlab-like environ- ment for machine learning},
	year = {2011}
}

@ARTICLE{2020SciPy-NMeth,
  author  = {Virtanen, Pauli and Gommers, Ralf and Oliphant, Travis E. and
            Haberland, Matt and Reddy, Tyler and Cournapeau, David and
            Burovski, Evgeni and Peterson, Pearu and Weckesser, Warren and
            Bright, Jonathan and {van der Walt}, St{\'e}fan J. and
            Brett, Matthew and Wilson, Joshua and Millman, K. Jarrod and
            Mayorov, Nikolay and Nelson, Andrew R. J. and Jones, Eric and
            Kern, Robert and Larson, Eric and Carey, C J and
            Polat, {\.I}lhan and Feng, Yu and Moore, Eric W. and
            {VanderPlas}, Jake and Laxalde, Denis and Perktold, Josef and
            Cimrman, Robert and Henriksen, Ian and Quintero, E. A. and
            Harris, Charles R. and Archibald, Anne M. and
            Ribeiro, Ant{\^o}nio H. and Pedregosa, Fabian and
            {van Mulbregt}, Paul and {SciPy 1.0 Contributors}},
  title   = {{{SciPy} 1.0: Fundamental Algorithms for Scientific
            Computing in Python}},
  journal = {Nature Methods},
  year    = {2020},
  volume  = {17},
  pages   = {261--272},
  adsurl  = {https://rdcu.be/b08Wh},
  doi     = {10.1038/s41592-019-0686-2},
}

\end{document}